\documentclass[11pt,a4paper]{article}
\usepackage[hyperref]{acl2021}
\usepackage{times}
\usepackage{latexsym}
\usepackage{multicol}
\usepackage{multirow}

\usepackage{listings}
\usepackage{graphicx}
\usepackage{graphics}
\usepackage{soul}

\usepackage{microtype}

\aclfinalcopy 

\title{SemEval 2022 Task 12:  Symlink\\
Linking Mathematical Symbols to their Descriptions}

\author{$^1$Viet Dac Lai, $^1$Amir Pouran Ben Veyseh, $^2$Franck Dernoncourt, $^1$Thien Huu Nguyen \\ 
$^1$Dept. of Computer and Information Science, University of Oregon, Eugene, Oregon, USA\\
\textit{\{vietl,apouranb,thien\}@cs.uoregon.edu }\\
$^2$Adobe Research, Seattle, Washington, USA \\
\textit{franck.dernoncourt@adobe.com}
}

\date{}

\begin{document}
\maketitle
\begin{abstract}
We describe Symlink, a SemEval shared task of extracting mathematical symbols and their descriptions from LaTeX source of scientific documents. This is a new task in SemEval 2022, which attracted 180 individual registrations and 59 final submissions from 7 participant teams. We expect the data developed for this task and the findings reported to be valuable for the scientific knowledge extraction and automated knowledge base construction communities. The data used in this task is publicly accessible at \url{https://github.com/nlp-uoregon/symlink}.
\end{abstract}

\section{Introduction}

The exponential growth of published articles may exceeds many readers' ability to keep track of the development of their field of interest. Hence, automatic reading comprehension of scientific documents has attracted the attention of researchers across various domains such as Drug Discovery, Knowledge Base Construction, and Natural Language Processing. A crucial aspect of understanding scientific literature is understanding terminologies and formulae because they offer an explicit and precise interface to present the relation between scientific concepts \cite{schubotz2018improving}. As such, a reading comprehension machine needs to (i) identify their descriptions and formulae, (ii) segment them into primitive terms and symbols, and (iii) link the associated terms and corresponding symbols.

Working with mathematical formulae is arduous due to two fundamental reasons. First, common text encodings such as ASCII and Unicode do not fully support typing mathematical symbols.  As a result, complex mathematical formulae are rarely written using either ASCII or Unicode. Rather, a higher level encoding (or typesetting) is often used to encode the content of scientific documents, in particular LaTeX. Second, most scientific documents are stored in one of two forms: photos or Portable Document Format (PDF). Scientific documents that were published prior to the graphical computer era are printed and now scanned and distributed as photos. Nowadays, scientific documents are often composed in some text editors or word processing software, then exported and shared a PDF file. Unfortunately, analyzing textual information in photo images or PDF files is extremely difficult, and most of the natural language processing tools are not developed to handle this format. As such, to facilitate the understanding of scientific literature, documents should be stored using a universal easy-to-process text-like encoding. In this paper, we use LaTeX as the typesetting to facilitate document analysis. 
Thanks to recent advances in text processing and image recognition, a LaTeX document can often be restored to some extent from either a photo or a PDF file \cite{deng2017image}.

\begin{figure*}
    \centering
    \includegraphics[width=\textwidth]{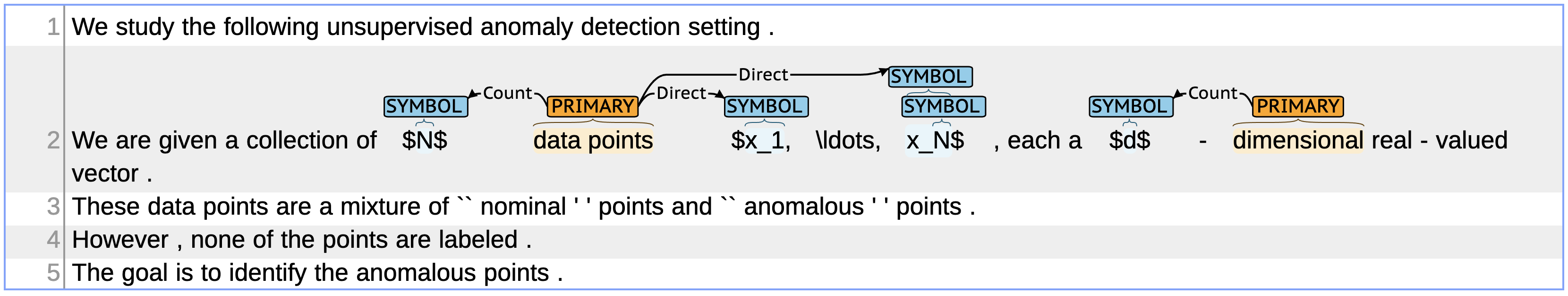}
    \caption{Example of the Symlink tasks.}
    \label{fig:task-example}
\end{figure*}

This paper introduces the Symlink shared task for the extraction of mathematical symbols and their descriptions from English scientific documents using their LaTeX source.  Figure \ref{fig:task-example} visualizes an example of the task. This paper also presents an analysis of the results of participant systems on the task. The rest of the paper is organized as follows. Section \ref{sec:related-work} presents related work in extracting formulae and their related information from scientific documents. Section \ref{sec:task} describes the subtasks of this Symlink shared task. Section \ref{sec:data} presents the data creation process including data sources, preprocessing, annotation guidelines, annotation, and data format. An analysis of the created data set is provided in Section \ref{sec:data-analysis}. The evaluation method is presented in Section \ref{sec:evaluation}, while the descriptions of the submitted systems are presented in Section \ref{sec:system}.

\section{Related Work}
\label{sec:related-work}

Early studies for scientific literature link formulae to Wikipedia page \cite{nghiem2010mining,kristianto2016entity}. Even though this can provide additional information regarding the mathematical expression, a reader might find it harder to understand the Wikipedia page as it is presented in many unrelated forms. Linking to the description in the same document is more practical \cite{kristianto2014extracting,alexeeva2020mathalign} as the descriptions are dedicated to the symbols and the context presented in the document. 

Previous studies on symbol-description extraction rely on pattern matching \cite{yokoi2011contextual,nghiem2010mining} and rule-based algorithms \cite{alexeeva2020mathalign}. These methods might work for observed patterns with an assumption of close proximity between symbol and description. They may fail to capture distant symbol-description pairs and symbols in very complex structures such as algorithms in computer science literature. 

Most of the previous studies have attempted to extract and link at formula level \cite{nghiem2010mining,kristianto2014extracting, kristianto2016entity}. In reality, understanding mathematical formulae requires details of atomic symbols e.g. superscript, subscript, function arguments. We believe that addressing the problem at this fine-grain level is crucial to drive future research toward a better understanding of the complex symbol-description extraction task.

Prior to this shared task, some studies have created datasets for similar tasks \cite{yokoi2011contextual,schubotz2016semantification,alexeeva2020mathalign}. However, one of them is created for publications written in Japanese \cite{yokoi2011contextual}, making it nearly impossible to transfer to English literature. While two other datasets \cite{schubotz2016semantification,alexeeva2020mathalign} only annotate small-scale golden datasets for evaluation purposes. As the result, no training data is available for training deep neural network models. In this shared task, we provide a large-scale dataset for English literature that we believe will provide enough supervision for the promising deep neural network-based models.

Definition extraction from scientific document is close to the task presented in SemEval Task 12. The Scientific Document Understanding workshop has hosted the Acronym Extraction and Acronym Disambiguation Shared Tasks, namely \textit{Acronym Extraction and Acronym Disambiguation Shared Tasks}\cite{SDUtaskpaper2021,SDUtaskpaper2022}. The prior studies in this research direction considers extracting definitions from the text  \cite{spala-etal-2019-deft,spala-etal-2020-semeval,veyseh2020joint}, or together with acronyms, and acronyms sense disambiguation \cite{pouran-ben-veyseh-etal-2020-acronym,pouran-ben-veyseh-etal-2021-maddog}.

\section{Task Description}
\label{sec:task}

The ultimate goal of Symlink shared task is to extract pairs of mathematical symbols and descriptions from scientific documents. As such, Symlink shared task is a combination of an entity recognition and an entity linking task.

Given a LaTeX source of a paragraph from a scientific document:

\begin{itemize}
    \item \textbf{Named Entity Recognition:} For each paragraph, identify all spans containing mathematical symbols and terminology descriptions. 
    
    \item \textbf{Relation Extraction:} For each pair of entities, identify the relationships between them if it is available among symbols and descriptions using Coref-Description, Coref-Symbol, Direct, Count relation types.
\end{itemize}


\section{Data Annotation}
\label{sec:data}


\subsection{Data source} 

We obtain the documents from \url{arXiv.org}, a repository for preprint scientific articles due to the broad coverage of subjects in scientific articles published in ArXiv. In particular, ArXiv offers articles in physics, mathematics, quantitative biology, computer science, quantitative finance, statistics, electrical engineering, and economics. As such, our obtained papers contain a large number of mathematical symbols and equations, allowing a higher yield of extracted symbol-description relations. Among these subjects, we choose five subjects of mathematics, physics, biology, economics, and computer science for annotation.

\subsection{Data preparation}

ArXiv open-sources the LaTeX version of their articles, when available. In order to make our  Symlink dataset open-access to the whole community, we crawled the metadata of these articles and only selected articles under the CC BY license. Once obtained the LaTeX project, we extracted all the paragraphs from the \textbf{.tex} files. We filtered out all short paragraphs with less than 50 words and paragraphs without symbols. Since a formula can be composed in multiple ways such as inline formulae (between $\$\;\;\$$), displayed formulae (between $\$\$\;\;\$\$$), or using commands e.g. \textit{array}, to keep the original TeX format of the formulae, all of these math objects are masked before tokenization. Then, we used the SciBERT tokenizer \cite{beltagy-etal-2019-scibert} to tokenize the text. The original math object is then restored. As we observed that many papers have nested math objects, we deleted all the nested objects, hence, having non-nested LaTeX data. This is helpful as it makes the LaTeX documents more similar to the ones generated by the PDF-to-LaTeX tools, which do not contain nested objects.

\subsection{Taxonomy}

To prepare for the annotation, we designed a taxonomy with 3 general entity types and four relation types. In particular, mathematical symbols are annotated under the tag \textbf{SYMBOL}, whereas descriptions are tagged under two labels \textbf{PRIMARY}, for single standalone definitions, and \textbf{ORDERED}, for the description of multiple terms, whose mentions are not separated without creating non-contiguous mentions. Due to the quadratic numbers of combinations of descriptions and complex math expressions, we only tagged an entity if and only if there is a second entity that pairs with the first entity to form a relationship. For relation, we are particularly interested in two main types of relations: \textbf{DIRECT}, linking a symbol with its definition, and \textbf{COUNT}, linking a description of a concept with a symbol that is the number of instances of the concept. Due to the sheer number of repetitions and coreferences of both descriptions and symbols, we also annotated \textbf{COREF-SYMBOL} relation, linking co-referred symbols, and \textbf{COREF-DESCRIPTION} relation, linking co-referred descriptions. Detailed annotation guidelines with examples are presented in Appendix \ref{app:guidelines}.

\subsection{Annotation}

We recruited 10 annotators from the crowdsourcing platform \url{upwork.com} to annotate scientific papers in the five mentioned domains (each subject was annotated by two annotators). The annotators are explicitly selected based on their demonstrated experiences in reading and writing scientific documents in their expertise field(e.g., holding an M.S. or Ph.D. degree). Detailed annotation guidelines with many examples and explanations are provided to train the annotators. Overall, we annotated 102 papers, accounting for 3,690 paragraphs, and 595K tokens. Our annotators for each domain co-annotate the documents in their domain and achieve Cohen's Kappa scores of (averaged) 0.79. This inter-agreement score thus indicates substantial agreements between our annotators. Eventually, the annotators engage in discussions to resolve any conflict to produce a final consolidated version of our Symlink dataset. 

\subsection{Data Format}

The participants are provided with preprocessed in JSON format. Each paragraph is stored in a JSON object with its id, topic, original LaTeX source, set of entities, and set of relations. An example of the data object is presented in Figure \ref{fig:data-example}.

\begin{figure}[t]
\begin{lstlisting}
{
  "id": "1503.01158v2...",
  "phase": "test",
  "topic": "cs.ai",
  "document": "1503.01158v2...",
  "paragraph": "paragraph_48",
  "text": "... with a covariance 
  matrix of $I$ ; that is , ...",
  "entity": {
   "T1": {
    "eid": "T1",
    "label": "SYMBOL",
    "start": 325,
    "end": 326,
    "text": "I"
   },
   "T2": {
    "eid": "T2",
    "label": "PRIMARY",
    "start": 303,
    "end": 320,
    "text": "covariance matrix"
   }
  },
  "relation": {
   "R1": {
    "rid": "R1",
    "label": "Direct",
    "arg0": "T2",
    "arg1": "T1"
   }
  }
}
\end{lstlisting}
\caption{An example of a paragraph in Symlink dataset.}
\label{fig:data-example}
\end{figure}

\section{Data Analysis}
\label{sec:data-analysis}

\begin{table}[t]
\caption{Statistics and label distribution of the Symlink dataset. $^*$The texts are tokenized by SciBERT.}
\resizebox{\linewidth}{!}{
\begin{tabular}{|l|r|r|r|r|}
    \hline
     & \textbf{Train} & \textbf{Dev} & \textbf{Test} & \textbf{Total}\\ \hline 
     \multicolumn{5}{|l|}{\textbf{Statistics}} \\
     \hline 
    \#Documents  & 91 & 6 & 5 & 102 \\
    \#Paragraphs & 3,120 & 270 &  300 & 3,690\\
    \#Sentences  & 25,070 & 1,765 & 2,286 & 29,121\\
    \#Tokens$*$ & 522K & 35K & 38K & 595K \\
    \hline
    \multicolumn{5}{|l|}{\textbf{Entity types}} \\
    \hline
    \#SYMBOL & 18,547 & 1,504 & 1,864 & 21,915\\
    \#PRIMARY & 7,953 & 678 & 907 &  9,538\\
    \#ORDERED & 14 & 3 & 1 & 18\\
    \hline
    \multicolumn{5}{|l|}{\textbf{Relation types}} \\
    \hline
    \#Direct            & 8,200 & 731    & 867   & 9,798\\
    \#Count             & 1,484 & 17     & 221   & 1,722\\
    \#Coref-Symbol      & 6,821 & 759    & 690   & 8,270 \\
    \#Coref-Description & 612  & 97     & 154   & 863 \\
    
    \hline
\end{tabular}
}
\label{tab:statistics}
\end{table}

Table \ref{tab:statistics} presents the statistics for the dataset including the number of articles, distribution of entities, and distribution of the relations. Overall, our dataset offers more than 31K entities, 20K pairs of relations, which is one order of magnitude larger than existing datasets for a similar task. 

\begin{figure}[!ht]
    \centering
    \includegraphics[width=\linewidth]{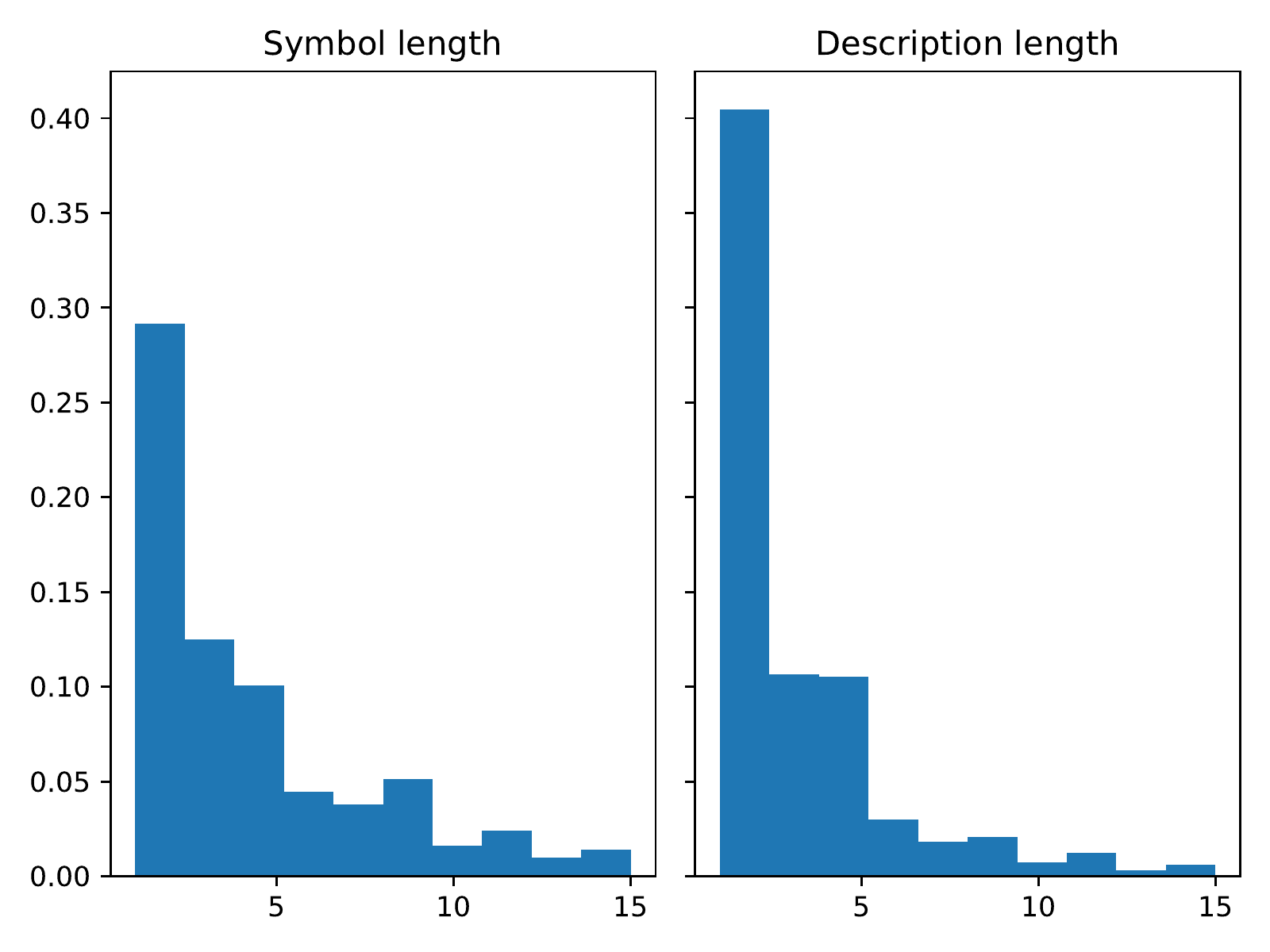}
    \caption{Length of symbols and descriptions in Symlink}
    \label{fig:span-length}
\end{figure}

Figure \ref{fig:span-length} presents the distribution of the span lengths of both symbols and descriptions of up to 15 tokens. As can be seen from the figure, the majority of entities have a length of 1-3 tokens. However, overall, the span lengths of both symbols and descriptions vary significantly from 1 up to 47 tokens (note that Figure \ref{fig:span-length} only illustrates the spans with up to 15 tokens). 
This demonstrates a key challenge of the Symbol-Description Linking task in this paper where symbols and descriptions with long spans might introduce confusion for extraction models.


\begin{figure}[!ht]
    \centering
    \includegraphics[width=\linewidth]{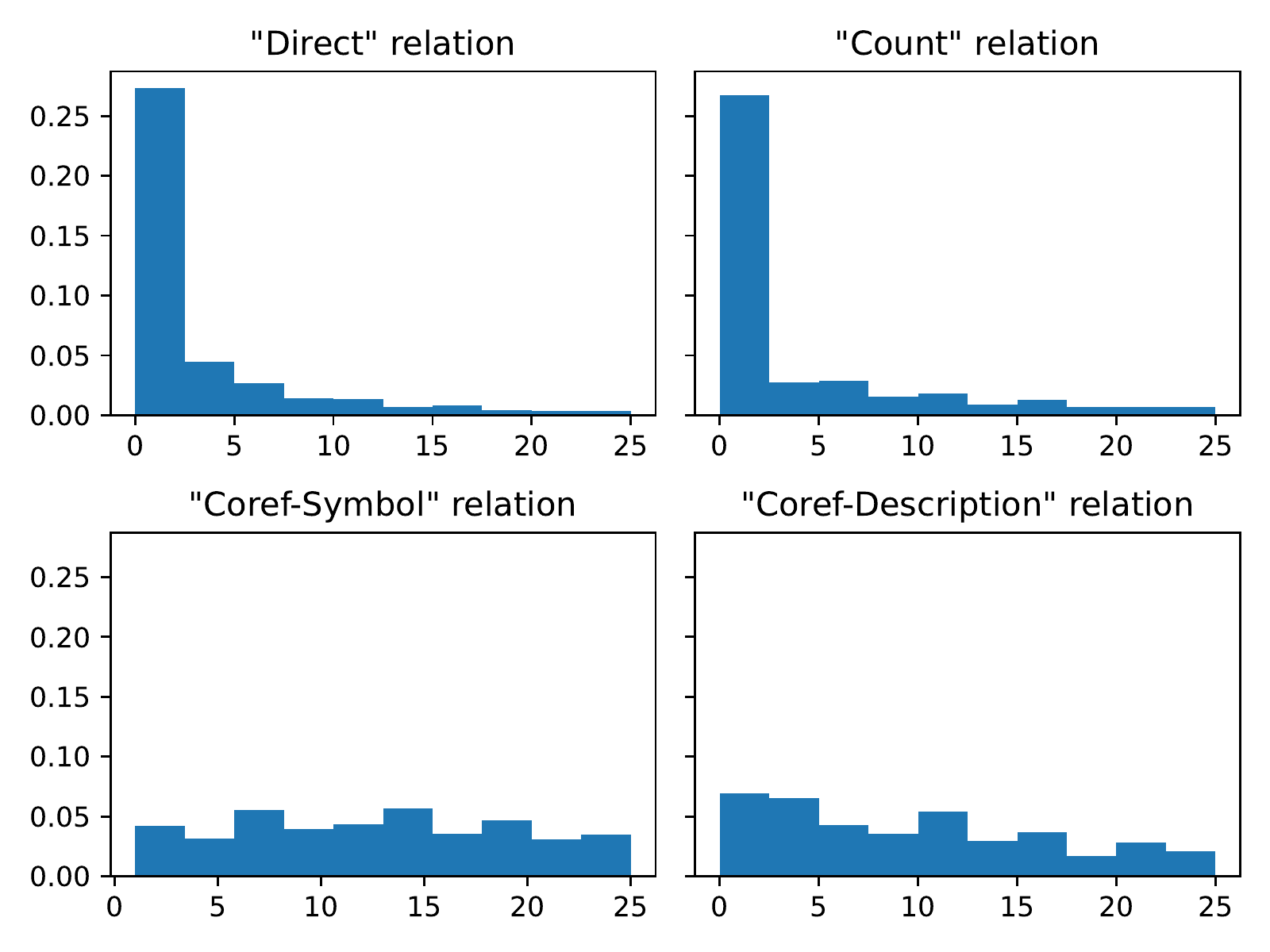}
    \caption{Distribution of distances between entities in Symlink by relation type.}
    \label{fig:distance}
\end{figure}

To further understand the dataset, we present the distances between the entities and relations annotated in Symlink by different relation types in Figure  \ref{fig:distance}. The distributions can be grouped into two categories. The first category involves the symbol-description relations while the second group involves the coreference relations. The distributions of symbol-description relations have long tails, indicating that symbols and descriptions tend to appear in close proximity. On the other hand, the distributions of coreference relations are quite flat, suggesting that the coreference relations appear in both short and long distances.

\section{Evaluation}
\label{sec:evaluation}

The results are evaluated separately for the Named Entity Recognition (NER) task and the Relation Extraction (RE) task. For NER, we use the entity-based partial/type from SemEval 2013 Task 9.1. For RE, we use standard precision, recall, F-score metrics. Relations output by the participating system is correct if the prediction label strictly matches the gold standard.

During the 21-day evaluation period (January 10 through 31, 2022), 7 CodaLab users submitted a total of 59 submissions with 37 submissions passing the validation and being scored. Given the complexity of the task, we allow unlimited submissions during the evaluation. As such the top submitter tried up to 18 times.

Table \ref{tab:result} shows the performances of the successful submissions. Asterisk denotes teams with system descriptions submitted for review. Among the participated teams, 6 teams performs both Named Entity Recognition and Relation Extraction subtasks while one team tried the Named Entity Recognition subtask only. Figure \ref{fig:submission} presents the timelines of submissions and high scores over the evaluation period.

\newcommand{\mytitle}[2]{\multicolumn{#1}{|c|}{\textbf{#2}}}

\begin{table*}[]
    \centering
    \caption{Results for each team/user, ordered by F1-score on Relation Extraction. Team with $*$ submitted their system description paper to SemEval 2022.}
    \begin{tabular}{|c|c|r|r|r|r|r|}
        \hline
        \multirow{2}{*}{\textbf{Team}}&  \multirow{2}{*}{\textbf{Variant}} & \mytitle{2}{Entity} & \mytitle{3}{Relation }\\
        \cline{3-7}
        & & \mytitle{1}{F1 (partial)} & \mytitle{1}{F1(type)} & \mytitle{1}{Precision} & \mytitle{1}{Recall} & \mytitle{1}{F-score} \\
        \hline
        \multirow{4}{*}{JBNU-CCLab*} & Base & 47.61 & 47.70 & 32.09 & 38.56 & 35.03  \\
                   & +RDrop & 47.61 & 47.70 & 33.40 & 38.66 & 35.84 \\
                   & +R3F   & 47.61 & 47.70 & 33.77 & 38.56 & 36.00 \\
                   & +R3F,Ensemble &\textbf{47.61} & \textbf{47.70} & 38.20 & \textbf{36.23} & \textbf{37.19} \\
       \hline
       ZQ & - & 39.39 & 39.51 & \textbf{57.25} & 23.29 & 33.11 \\
        \hline
       \multirow{4}{*}{AIFB-WebScience*} & Max/Original  & 37.83 & 37.88 & 45.80 & 20.96 & 28.66 \\
               & Mean/Original & 41.21 & 41.23 & 42.25 & 26.55 & 32.28 \\
               & Max/LaTex2Text & 38.33 & 38.38 & 46.09 & 21.64 & 29.45 \\
               & Mean/LaTex2Text & 34.53 & 34.64 & 47.02 & 18.20 & 26.24 \\
       \hline
       LingZing & - & 33.87 & 33.93 & 13.45 & 10.92 & 12.05 \\
        \hline
       \multirow{2}{*}{MaChAmp*} & Single mBERT & - & - & -  & -  & 2.67 \\
       & Multi RemBERT & 25.17 & 25.25 & 13.11 & 5.17 & 7.42 \\
       \hline
       iyerke & - & 6.67 & 6.46 & 0.10 & 0.62 & 0.17 \\
       \hline
       AN(L)P* & - & - & 16.30 & - & - & - \\
       \hline
    \end{tabular}
    
    \label{tab:result}
\end{table*}

\section{Summary of Participating Systems}
\label{sec:system}

The Symlink track at SemEval-2022 received 4 system description paper submissions presented in Table \ref{tab:result}. Overall, all submitted systems are based on BERT architecture \cite{devlin-etal-2019-bert}. Among those, two out of four systems use SciBERT \cite{beltagy-etal-2019-scibert}, while two remaining systems use other variants of BERT such as original BERT \cite{devlin-etal-2019-bert} and mBERT \cite{devlin-etal-2019-bert}. 

\subsection{System Specifics}

\citet{jbnu2022lee} (JBNU-CCLab) achieved their state-of-the-art performance using SciBERT \cite{beltagy-etal-2019-scibert}. Their entity model consists of an MRC-based model \cite{li-etal-2020-unified}, simplifying the tasks as binary classification problems whether span is valid using entity type information as input features. They proposed a simple rule-based  Symbol Tokenizer to predict accurately the complex symbols appearing in scientific documents. The relation model exploits entity span information and entity type information as input features using typed entity marker. Additionally, the paper exploited many regularization techniques to improve the model performance such as regularized dropout \cite{wu2021r} and representational collapse prevention \cite{aghajanyan2020better} and traditional ensemble techniques.

\citet{aifb2022popovic} (AIFB-WebScience) proposed an end-to-end joint entity and relation extraction approach based on transformer-based language models. Unlike traditional entity and relation extraction methods, which perform the task in sequence, this system incorporates information from relation extraction into entity extraction. As such, the system can be trained even on partially annotated datasets where only a subset of all valid entity spans is annotated. 

\citet{anlp2022ping} (AN(L)P) participated in the Entity Extraction only. They finetuned a BERT-large model \cite{devlin-etal-2019-bert} for each domain. For cs.ai domain, they used data from cs.ai only, whereas, for the other domain, they augmented the in-domain data with the data from cs.ai.

\citet{machamp2022goot} (MaChAmp) proposed to pretrain a language model and re-finetune after multi-task learning for a pre-defined set of semantically focused NLP tasks. They trained a multi-task model for all text-based SemEval tasks that include annotation on the word, sentence, or paragraph level. They compared the performance with models using mBERT \cite{devlin-etal-2019-bert}. The pretrained multi-task embedding showed a consistent improvement across many tasks against the mBERT embedding.

\subsection{Symbol tokenizer and detection}

In this shared task, the uniqueness of the task is detecting mathematical symbol span. Symbol span in LaTeX source is comprised of both human language and machine language, i.e. LaTeX language. Further, mathematical formulae in LaTeX sources are written in both linear and hierarchical manners. Therefore, a system must consider not only human language modeling but also a highly systematic syntax system of LaTeX source. As such, fundamental tasks such as tokenization is a huge contributor to the robustness of the model.

Among four submitted systems, MaChAmp \cite{machamp2022goot} and AN(L)P \cite{anlp2022ping} teams used the default tokenizer from either BERT or mBERT, which are not designed for scientific documents. Consequently, they are unable to correctly segment the mathematic source, hence, they achieved the lowest Named Entity Recognition performance. Whereas AIFB-WebScience \cite{aifb2022popovic} and JBNU-CCLab \cite{jbnu2022lee} achieved much higher performances thanks to SciBERT tokenizer because it is trained on scientific literature. However, the SciBERT tokenizer is far from perfect such  that JBNU-CCLab further proposed to tokenize the mathematical formulae using a customized rule-based tokenizer based on capital letters, numbers, and special
characters(e.g. \%, \$, \{, \}). Hence, they achieved state-of-the-art performance on both NER and RE subtasks.

\section{Conclusion}

In this paper, we present the task description, the data annotation, the evaluation, the results, and the descriptions of four submitted systems for Symlink at SemEval 2022. The Symlink shared task is challenging given the complexity of the LaTeX source and partly due to the difference of the domains involved in the data. In this shared task, it is hard to separate the NER and RE subtasks due to their constraints. 

The submitted systems employed variants of contextualized embedding BERT for encoding the text. In general, the task can be formatted into similar sequence labeling and relation extraction task. However, special treatments are needed to process LaTeX sources. For instance, a LaTeX-source-trained tokenizer or a customized tokenizer is essential to tokenize the text. Some unique characteristics of the dataset have not been investigated such as the syntax of the LaTeX source, and the hierarchical structure of formulae. These suggest future research directions to improve the robustness of the model. 


\bibliographystyle{acl_natbib}
\bibliography{anthology,acl2021}

\appendix
\section{Annotation guidelines}
\label{app:guidelines}

This section summarizes some rules that we use to make our annotations more consistent. 

\textbf{Description tagging}: A description is usually a noun or a noun phrase that expresses a concept. These are the overall rules for entity annotations:
\begin{itemize}
    \item We only tag a description if the corresponding symbol presents in the text.
    \item A description usually is a noun or a noun phrase. Sometimes, a verb, an adverb, or an adjective describes an operation, it is also considered a description.
    \item Descriptions should be short but it must cover the elements in the corresponding symbol, esp. in case of complex symbols, such as superscript, subscript, arguments, and limits.
\end{itemize}

\textbf{Symbol tagging}: A mathematical symbol can present an operand, an operator, an expression, or combination of these.
\begin{itemize}
    \item An atomic symbol in PDF format has to be a character, that means, if  we have Y hat, neither Y nor hat is considered an atomic symbol, instead “Y hat” is a symbol. In latex format, \textbackslash hat\{Y\}  should be annotated.
    \item A complex symbol is a combination of multiple symbols and brackets, for example: “P(x)”, “Wx”
    \item An annotated symbol has to be a complete symbol e.g. “P(x)” is good, “P(x” is not because of lacking the closing parenthesis.
    \item A complex formula can be segmented into atomic symbols, we will annotate at all levels of the complex symbol as long as there are appropriate descriptions available.
\end{itemize}

\textbf{Relation annotation}: 

\begin{itemize} 
    \item Every annotated symbol/description has to have at least one relation linking to its description/symbol. 
    \item If there are multiple mentions of a single symbol/description, use coreference relation to link them. A direct relation or a count relation is used to link the closet pair of symbol and description.
\end{itemize}

\section{Timeline of submissions}

\begin{figure}
    \centering
    \includegraphics[width=\linewidth]{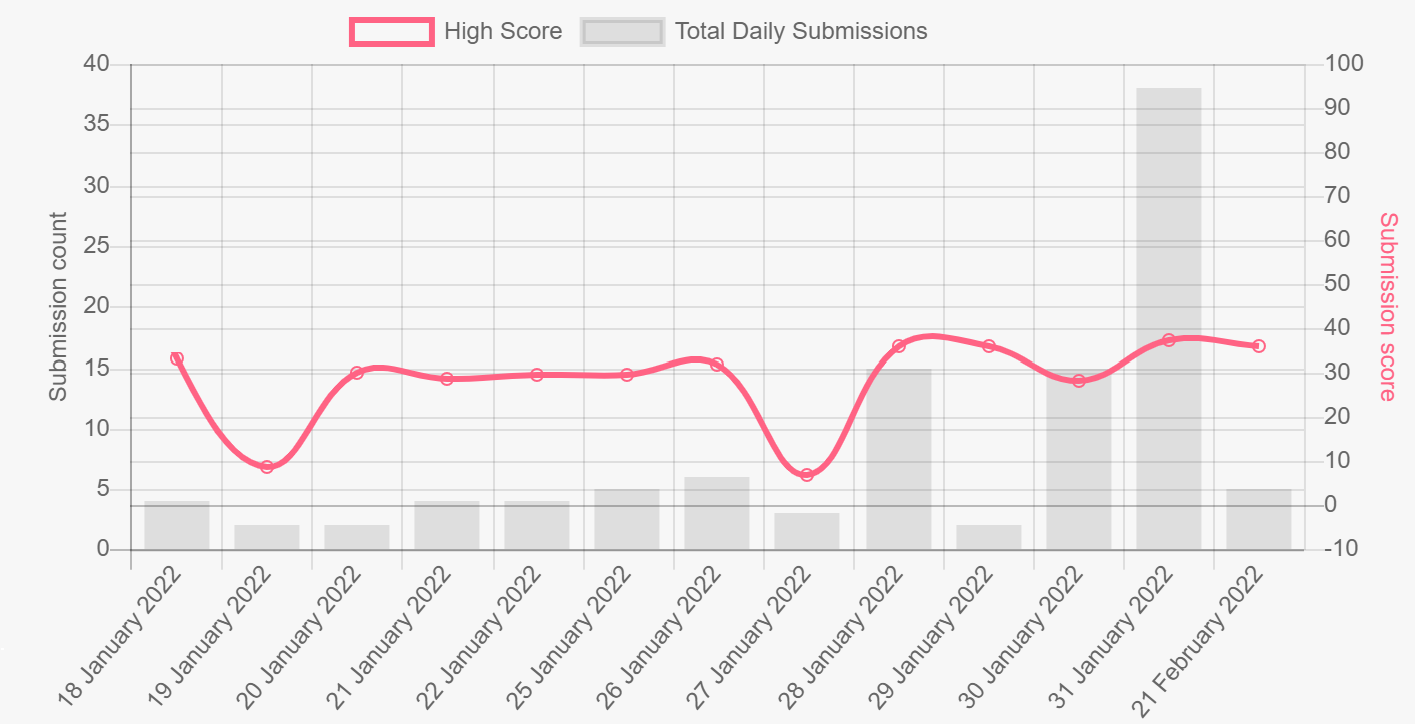}
    \caption{Submission counts and top performances during the evaluation period. The submission score is the F1-score of the RE task.}
    \label{fig:submission}
\end{figure}

Figure \ref{fig:submission} presents the number of submissions over the evaluation of the task.

\end{document}